\pdfoutput=1

\documentclass[11pt]{article}

\usepackage{acl}

\usepackage{times}
\usepackage{latexsym}

\usepackage[T1]{fontenc}
\usepackage{ragged2e}
\usepackage{siunitx}
\usepackage{adjustbox}
\usepackage{amsmath,amsfonts,bm}
\usepackage{mathtools}
\newcommand*\rot{\rotatebox{90}}
\newcommand{\STAB}[1]{\begin{tabular}{@{}c@{}}\rot{#1}\end{tabular}}

\usepackage{amsfonts}

\usepackage[utf8]{inputenc}

\usepackage{microtype}

\usepackage[subtle]{savetrees}
\usepackage{multirow}
\usepackage{hyperref}
\usepackage{booktabs} 
\usepackage{tabularx}
\usepackage{graphicx}
\usepackage{subcaption}
\usepackage[htt]{hyphenat}
\usepackage{sidecap}

\usepackage{placeins}

\makeatletter
\newcommand{\mathleft}{\@fleqntrue\@mathmargin0pt}
\newcommand{\mathcenter}{\@fleqnfalse}
\makeatother
\usepackage{titlesec}
\titlespacing*{\section}
{0.5ex}{0.5ex}{0.5ex}
\titlespacing*{\subsection}
{0.5ex}{.5ex}{.5ex}
\titlespacing*{\subsubsection}
{0pt}{.7ex}{.7ex}

\title{Mix and Match: Learning-free Controllable Text Generation\\ using Energy Language Models \vspace{0ex}}

\author{Fatemehsadat Mireshghallah\textsuperscript{\rm 1}, Kartik Goyal\textsuperscript{\rm 2},
    \textbf{Taylor Berg-Kirkpatrick}\textsuperscript{\rm 1} \\
    \textsuperscript{\rm 1} University of California San Diego,
    \textsuperscript{\rm 2} Toyota Technological Institute at Chicago (TTIC)\\
    \texttt{[fatemeh, tberg]@ucsd.edu}, \texttt{ kartikgo@ttic.edu}\\
  }

\begin{document}
\maketitle

\begin{abstract}
\vspace{-1.7ex}

Recent work on controlled text generation has either required attribute-based fine-tuning of the base language model (LM), or has restricted the parameterization of the attribute discriminator to be compatible with the base autoregressive LM. In this work, we propose Mix and Match LM, a global score-based alternative for controllable text generation that combines arbitrary pre-trained black-box models for achieving the desired attributes in the generated text without involving any fine-tuning or structural assumptions about the black-box models. We interpret the task of controllable generation as drawing samples from an energy-based model whose energy values are a linear combination of scores from black-box models that are separately responsible for fluency, the control attribute, and faithfulness to any conditioning context. We use a Metropolis-Hastings sampling scheme to sample from this energy-based model using bidirectional context and global attribute features. We validate the effectiveness of our approach on various controlled generation and style-based text revision tasks by outperforming recently proposed methods that involve extra training, fine-tuning, or restrictive assumptions over the form of models.   

\end{abstract}

\section{Introduction}
 \begin{figure*}[h!]
    \centering
     \includegraphics[width=0.8\textwidth]{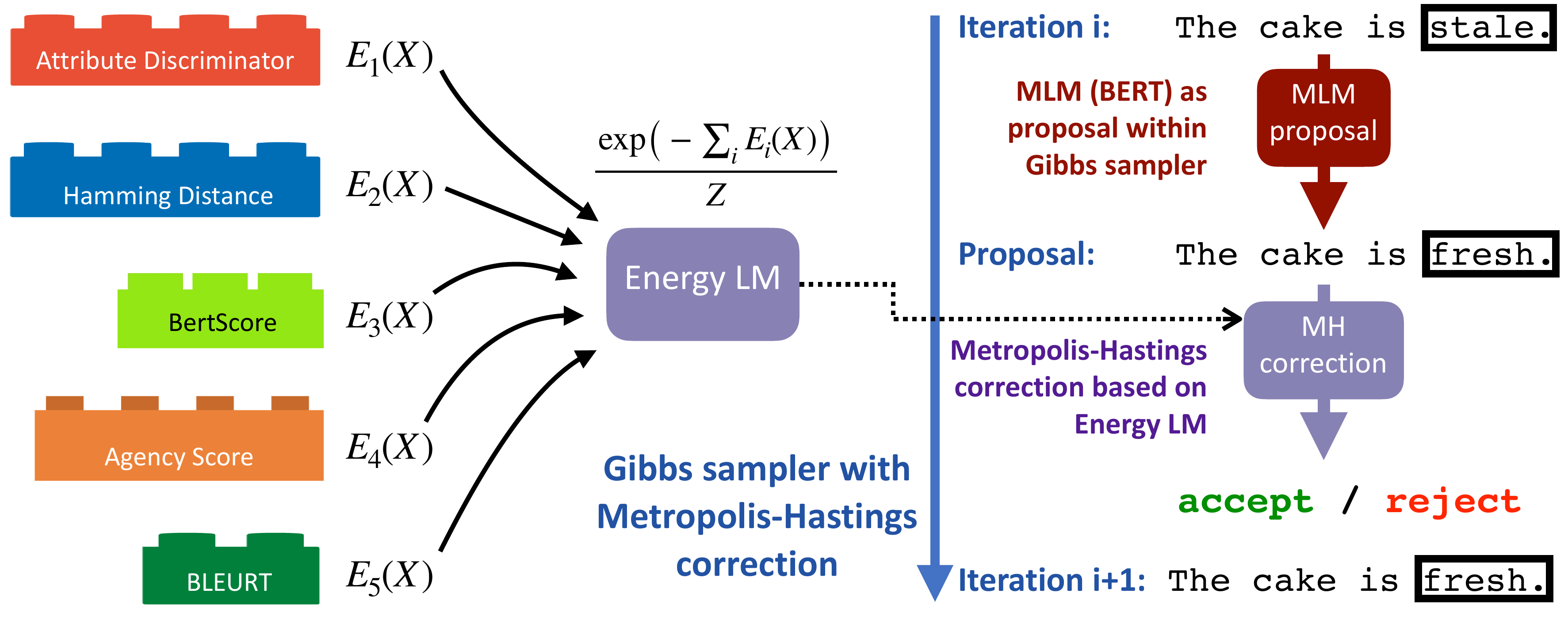}
     \caption{Overview of Mix and Match LM. The Lego pieces show different experts that can be used to form the energy LM and help control different features in the generated text. The right side shows the $i$th step in the the Gibbs sampling chain, where a proposal is made by the MLM, and then it is accepted/rejected based on the energy score.}
     \label{fig:mmlm}
    \vspace{-1ex}
 \end{figure*}

While large transformer-based autoregressive language models trained on massive amounts of data found on the internet exhibit exceptional capabilities to generate natural language text, effective methods for generating text that satisfy global constraints and possess holistic desired attributes remains an active area of research. These mechanisms for controlling the generation of language have the potential to mitigate undesirable biases encoded by the large language models and prevent the generation of hate speech and toxic language~\cite{xudetoxifying,gehman2020realtoxicityprompts,sap2021annotators,baheti2021just,mireshghallah-berg-kirkpatrick-2021-style}.
Much of the prior work has approached controlled generation via either training  domain-conditioned neural language models~\cite{prabhumoye-etal-2020-exploring, he2020, lample2018phrase,shen2017style,Krishna2020ReformulatingUS, reif2021recipe, ficler-goldberg-2017-controlling, khalifa2021a} or finetuning/modifying an underlying large pre-trained base model for generation on domain-specific data for attribute sensitive generation~\cite{ziegler2019fine, keskar2019ctrl,mai-etal-2020-plug,gururangan-etal-2020-dont, chronopoulou2021efficient}. Not only do these approaches involve computational overhead and estimation errors associated with the training of language models, but they are also dependent on access to a large amount of attribute-specific language data which can be impractical in many scenarios and exacerbate privacy concerns~\cite{brown2022does,mireshghallah-etal-2021-privacy,Kandpal2022DeduplicatingTD}.

Our approach eschews training and focuses on generation-time control from pre-trained modules. Recent work in this space has used attribute discriminators \cite{pplm, gedi, fudge, holtzman-etal-2018-learning} to steer the generation from a large autoregressive language model. These discriminators need to be separately trained on partial generations in order to be operationalized with step-wise autoregressive models. As a result, this approach also requires availability of data to train step-wise discriminators for attributes that are essentially global (at the sequence-level) in nature. 
Therefore, we focus on drawing samples from a test-time combination of \emph{pretrained blackbox} experts that each score a desired property of output text -- for example, fluency, attribute sensitivity, or faithfulness to the context. Specifically, we view the product of these black-box experts as a probabilistic energy model~\cite{poe} -- i.e., a \emph{non-autoregressive}, globally normalized language model -- and then sample (without further training or fine-tuning) using a specialized Gibbs sampler with a Metropolis-Hastings correction step~\cite{MH}.

Our full framework, which we entitle Mix and Match LM (depicted in Figure~\ref{fig:mmlm}), enables the generation of high-quality attribute-controlled samples by mixing and matching black-box models like off-the-shelf pre-trained attribute-sensitive discriminators (e.g., sentiment classifiers), large bidirectional pre-trained language models like BERT~\cite{bert}, and other modules specializing in capturing desirable features pertaining to faithfulness to any additional context, like hamming distance, or BertScore distance~\cite{bertscore} between the sample and the conditioning context. 
We generate samples from the energy language model assembled from these component experts by using the recently proposed Gibbs-Metropolis-Hastings scheme~\cite{MH} for sampling from energy models using a masked language model as a proposal distribution. In this scheme, an expressive bidirectional language model like BERT is used to make a proposal at each transition step in the Gibbs chain to jump to a sequence $\mathbf{\bar{x}}$ from the current sequence $\mathbf{x}$. This proposal's fitness is judged by the change in the energy language model's score, with the sampler accepting proposals with larger energy reductions at a higher rate. While the MCMC nature of our sampler negatively impacts the runtime during decoding compared to autoregressive approaches with ancestral sampling, we find our approach to still be practical and yield high-quality diverse samples that respect the distribution induced by the product of expert black-box models. 

We demonstrate the flexibility of our approach by performing a variety of controlled generation tasks, such as aspect-based text revision, style transfer, and attribute grounded generation and compare it to recently proposed controlled generation approaches that are more resource/data intensive. We observe that our approach, which does not require any gradient optimization and is able to combine arbitrary heterogeneous black-box models, outperforms other approaches according to various automated metrics of fluency, quality, and control, as well as human evaluations. 
 We have provided code, data, and sample generations in this GitHub repository:~\url{https://github.com/mireshghallah/mixmatch}~(see~\ref{app:code} for details on reproducing the results). 


\section{Related Work} The approaches closest in spirit to our work involve steering generation from a base language model with external attribute-sensitive control mechanisms. Plug-and-Play LM~\cite{pplm} uses discriminators learned from an autoregressive LM's top-level hidden layer to modify the LM's states toward increasing the probability of the desired attribute via gradient ascent at each step.
GeDi~\cite{gedi} and FUDGE~\cite{fudge} take a similar approach but train custom step-wise attribute-sensitive discriminators that decide whether the desired attribute is likely to be satisfied by the current generation path. GeDi trains class-conditional language models for these discriminators and hence additionally relies on access to attribute sensitive language data. \citet{kumar2021controlled} formulate the task of controlled generation as optimizing the base LM's likelihood subject to global differentiable attribute-based constraints by gradient descent over the position-wise simplexes over the vocabulary. DExperts~\cite{liu-etal-2021-dexperts} is another decoding-time controllable generation approach that modifies the step-wise softmax logits of an autoregressive pre-trained LM with softmax logits of separately trained domain-specific \emph{expert} autoregressive language models.
These approaches require training of custom modules and do not readily enjoy the benefits of incorporating global attribute-based features into the generation mechanism in a simple probabilistic manner.
In contrast, following the findings related to implicit energy-based models trained via non-probabilistic objectives \cite{grathwohl2019your, MH}, our energy-based formulation is not only optimization-free but also fully modular and able to easily incorporate global features, allowing for heterogeneous black-box experts to be combined with each other.

\section{Mix-and-match Language Models}
In this section, we describe our approach and motivation behind our method. Specifically, we frame the problem of performing controlled generation as a problem of sampling from a specialized energy-based (or globally normalized) sequence model that defines a probability distribution that satisfies the desired constraints we wish to impose in the controlled generation setting. As described below, this energy-based model is composed of pre-trained components and does not require any further optimization.  An energy-based sequence model defines the probability distribution over the space of possible sequences $\mathcal{X}$ as:\footnote{For simplicity, we are concerned with a finite set of sequences limited by some maximum length.} $p(X;\theta) = \frac{e^{-E(X;\theta)}}{\sum_{X' \in \mathcal{X}} e^{-E(X'; \theta)}}$,
where $E(X;\theta)$ refers to the scalar energy of a sequence $X$ that is parametrized by $\theta$. Lower energy corresponds to the higher likelihood of $X$. In contrast to the common autoregressive sequence models, exact likelihood computation and efficient sampling from these models is challenging. Despite these challenges, we focus on this paradigm of sequence modeling because energy-based models offer increased flexibility via sequence-level features and constraints. As we discuss next, this capability lets us easily define expressive functions for controlled generation of sequences which is not readily offered by the autoregressive modeling paradigm.
\subsection{Product of Experts Energy-based Models and Controlled Generation}
Our approach is motivated by the perspective that the task of controlled generation requires concentrating probability mass over 
a small subspace of sequences in $\mathcal{X}$ that satisfies various constraints pertaining to fluency, target attributes, and other control variables. Consider the task of generating positive sentiment sentences. This requires satisfaction of two major constraints: (1) The sequence $X$ should be well-formed, (2) The sequence $X$ should express positive sentiment. If we have access to two separate probability distributions over $\mathcal{X}$, one for modeling well-formedness  ($p_1(X)$) and another for modeling positivity ($p_2(X)$), then a natural solution for controlled generation in this setting would be to draw samples from a probability distribution that is a product of these two distributions i.e.
$p_{\texttt{desire}}(X) \propto p_{1}(X) \cdot p_{2}(X)$.
In our approach, we further relax this requirement by assuming access to \emph{expert blackboxes} that yield scalar non-probabilistic energy scores $E_1$ and $E_2$ indicating fitness of a sequence w.r.t. well-formedness and positivity respectively. Under the product of experts framework above the desired probability distribution would take the form:
$\log~p_{\texttt{desire}}(X) = -(E_1(X) + E_2(X))~-~\log Z$.
This expression shows that when working with scalar scores for the expert black-boxes, the product of expert models yields an energy model whose energy is simply the sum of the scalar energy values obtained from the expert models. Inspired by this, we propose a framework for controlled generation that involves linear combinations of various black-box experts in order to obtain a distribution whose samples satisfy the requirements of a desired controlled generation task:
$E_{\texttt{M\&M}}(X) = \sum_{i=1}^k \alpha_i E_{i}(X)$,
where our proposed \emph{mix-and-match} energy is composed of $k$ expert energy components, which are weighted by scalar hyperparameters $\alpha$.

\subsection{Expert Factors in Mix-and-Match LM}
As shown in Fig.~\ref{fig:mmlm}, we use the following black-box experts in our experiments as modules that we can add or remove to produce desired behavior:

\noindent $\mathbf{E_{\texttt{mlm}}(X)}:$ Recent work has shown that large masked language models~(MLM) like BERT can discriminate between well-formed and ill-formed sentences \cite{bertscore} and induce an implicit energy function over the sequences \cite{MH}. Hence, we use BERT-base as a black-box to model the form and fluency of sentences. Specifically, we use an energy parametrization introduced in \citet{MH} which is negative of the sum of unnormalized logits iteratively computed at each position obtained via the forward pass of the MLM after masking the corresponding position. 

\noindent $\mathbf{{E_{\texttt{disc}}(X)}}:$ This particular expert module refers to the energy obtained via the discriminator for the attributes of interest. What this module returns is the raw logits of the discriminator, for the target attribute. For instance, if we have a sentiment classifier, and want to produce positive sentiment, then ${{E_{\texttt{disc}}(X)}=-\log~p(+|X)}$. 
\par
\noindent $\mathbf{E_{\texttt{hamm}}(X;X')}:$ For a given sequence $X'$, this quantity refers to the hamming distance between the sequence $X$ and $X'$. This penalizes token level deviation from $X'$ which is useful if we are interested in only making minor edits to $X'$ as described later.
\par 
\noindent $\mathbf{E_{\texttt{fuzzy}}(X;X')}:$ Similar to the hamming distance, this quantity refers to the BertScore \cite{bertscore} computed between $X$ and $X'$ which can be viewed as a \emph{fuzzy} hamming distance that takes semantic similarity into account.

\subsection{Sampling scheme}
To sample from the energy parametrizations described in the previous section, we follow the Metropolis-Hastings~\cite{MHorigin} MCMC scheme for sampling from the masked language models introduced by \citet{MH}. While the proposal distribution we use is the same as \citet{MH} i.e. masked language model's (BERT's) conditionals, the energy parametrizations we use are more suitably designed for controlled generation.

We briefly explain the sampling procedure, which involves forming long Markov chains of sequences starting with a random sequence, and following the MH scheme which uses a proposal distribution to propose a new sequence at each step in a chain which is either accepted or rejected based on its fitness to the energy function. The sequences at the end of these chains correspond to samples from the desired energy-based model. Operationally, at each MCMC step, we mask out a token at a random position in the current sequence $X$ in the chain and propose a new sequence $\bar{X}$ to transition to by sampling a token from the MLM conditional softmax at the masked position. This proposed sequence is evaluated by its ability to reduce the energy from the current sequence in the chain and is accepted with the probability $p(\bar{X};X) = \min\left( 1,\frac{e^{-E_{\texttt{M\&M}}(\bar{X})}~p_{\textrm{mlm}}(X_i \mid X_{\backslash i})}{e^{-E_{\texttt{M\&M}}(X)}~p_{\textrm{mlm}}(\bar{X}_i \mid X_{\backslash i})} \right)$. $E_{M\&M}(X)$ refers to the product of experts energy, $i$ refers to the position chosen for masking, $p_{\textrm{mlm}}$ refers to the MLM's conditional distribution at the \texttt{[MASK]} position. Intuitively, this acceptance probability indicates that the proposed sequence $\bar{X}$ is more acceptable if it has lower energy than the current sequence $X$ in the chain and is rare or less likely to be proposed by the proposal distribution again.

\subsection{Controlled generation Tasks}
\label{sec:tasks}
We use the expert black-box factors and the sampling scheme described above in our framework to perform two kinds of controlled generation tasks.
\par
\noindent \textbf{Prompted generation:} This task focuses on generating well-formed sentences that start with a specified prompt and also satisfy a target attribute for which we have access to a discriminator. An example task would be to generate positive sentiment sequences starting with \texttt{This} \texttt{movie}. The energy function takes the form:
\begin{equation}
\label{eq:disc}
\begin{split}
E_{\texttt{gen}}(X)= E_{\texttt{mlm}}(X)~+~\alpha~E_{\texttt{disc}}(X) 
\end{split}
\end{equation}

$\alpha$ is a hyperparameter that controls the tradeoff between the MLM score and the discriminator's influence. 
For MH-based sampling for this task, we initialize the sequence with the starting prompt and the rest of the tokens masked out, which creates a seed text of shape~\texttt{the \; movie [MASK] [MASK] ... [MASK]}, for the prompt example of \texttt{the \; movie}. The number of mask tokens depends on the target generation length, and we constrain the sampler to only produce proposals and revise non-prompt tokens, and mark the prompt tokens as ``frozen''.
\par
\noindent \textbf{Controlled text revision}: This task involves editing a source sequence $X'$ in order to satisfy the desired target attributes exhibited by the generated sequence $X$. The energy function for this task is:
\mathleft
\begin{equation}
\scriptstyle 
\begin{split}
\scriptstyle
\label{eq:exp}
 E_{\texttt{rev}}(X)= E_{\texttt{gen}}(X)+\beta~E_{\texttt{hamm}}(X,X')+\gamma~E_{\texttt{fuzzy}}(X,X')
\end{split} 
\end{equation}
This energy function in addition to valuing well-formedness and satisfying target attribute requirements also focuses on maintaining faithfulness to the source sequence $X'$.
For sampling with this energy, we initialize the sequence with the sequence $X'$ to be edited. This sets the length of the target sequence to be the same as the source. In this setup, the sampler can revise all tokens and is not constrained.

For both these tasks, we run a separate MCMC chain for each generated sentence for 8 to 15 epochs, depending on the task. An epoch refers to one masking cycle over all the non-frozen positions (selected randomly) of the sequence.

\section{Experimental Setup}
We provide full experimental details in appendix Section~\ref{app:exps}, here we provide a brief overview of the tasks, datasets, baselines, and metrics used in the experiments.  

\subsection{Tasks and Datasets}

\noindent\textbf{Controllable debiasing (ROC story corpus):} We use the subset of the ROC story corpus~\cite{mostafazadeh-etal-2016-corpus} test-set that is used by PowerTransformer~\cite{ma-etal-2020-powertransformer} for their evaluations. We use this data for controllable debiasing, a text revision task which aims to correct the implicit and potentially undesirable agency biases in character portrayals,  by replacing verbs such as ``wish" and ``dream", with ``pursue" and ``achieve".

\noindent\textbf{Sentiment transfer (Yelp):} We use Yelp~\cite{shen2017style} dataset's test-set for the task of sentiment transfer. The test set comprises 1000 sentences, half with positive and half with negative sentiment. We also have a reference set of handwritten sentiment transferred sentences, provided by~\cite{he2020} that we use for reporting evaluation metrics.

\noindent\textbf{Formality transfer (GYAFC):}  We use 1051 sentences from the entertainment and music domain subset of the  GYAFC~\cite{Rao2018DearSO} dataset, which contains formal and informal sentences for the task of formality transfer (both directions of formal to informal and informal to formal).

\noindent\textbf{Prompted generation:} We evaluate our approach on two forms of prompted generation: 1) sentiment controlled generation and 2) topic controlled generation. For sentiment controlled generation, we set Mix and Match LM to generate text with positive or negative sentiment given prompts, by using a Yelp sentiment classifier as discriminator and compare against PPLM~\cite{pplm} which is a popular sentiment controlled generation method. For topic controlled generation, we compare against FUDGE~\cite{fudge}, and follow their experimental setup consisting of 7 distinct topics and 20 prompts. 

\subsection{Expert Component Configurations}

We use a Huggingface pre-trained \texttt{bert-base-uncased} model as our MLM for yielding $E_{\texttt{mlm}}$ and also providing the proposal distribution in our MH MCMC sampler. For obtaining $E_{\texttt{disc}}$, we train BERT-based classifiers on the training-set of our datasets to use as our attribute discriminators.
We could have used any pre-trained attribute classifier from  Huggingface for $E_{\texttt{disc}}$, but we keep those aside to use as external attribute classifiers for fair evaluation against baselines. 
For experiments in which we add the BertScore~\cite{bertscore} component to the energy, we use the pre-trained \texttt{roberta-large\_L17} model.
Finally, for agency score, we use the lexicon provided by~\cite{sap-lexicon} and check each generated sequence and count the number of target agency verbs that exist there. The count becomes the agency score. 

\subsection{Baselines}

\noindent\textbf{PowerTransformer.}
For the task of controllable debiasing (agency revision), we compare our work with PowerTransformer~\cite{ma-etal-2020-powertransformer}, an approach that uses paraphrasing and self-supervision based on a reconstruction loss, building on pre-trained language models, to re-write text and control agency level of sentences.

\noindent\textbf{~\citeauthor{he2020}} For style transfer on sentiment an formality, we compare with~\citet{he2020}, a generative style transfer framework which uses a variational autoencoder (VAE) built using a sequence-to-sequence LSTM-based model to do unsupervised style transfer. This framework needs to be trained from scratch for each style transfer task.

\noindent\textbf{UNMT.} As a second baseline for style transfer, we use UNMT~\cite{lample2018phrase}, an unsupervised machine translation framework that demonstrates high performance for sentiment transfer.

\noindent\textbf{PPLM.} For the task of sentiment controlled generation, we compare to Plug-and-Play LM (PPLM)~\citet{pplm}, which does attribute controlled generation using the flow of gradients from discriminators trained on the last hidden layer representations of the generator, to guide generation. 

\noindent\textbf{FUDGE.} This approach~\cite{fudge} trains step-wise discriminators on partial generations from GPT-2 to determine whether the constraints related to desired attributes will be satisfied by the future completion of the sequence or not. We compare against this on topic controlled generation as this approach was shown to be superior to PPLM on this task.

\subsection{Evaluation Metrics}
We use a variety of evaluation metrics to compare our approach's performance on two major facets: (1) Quality of generated text, and (2) success on matching the target attribute used for control.

\subsubsection{Text Quality and Semantic Similarity}
\label{sec:metric:lang}
\noindent\textbf{GPT-2 PPL.}  We feed our generated test sentences to a Huggingface~\cite{Radford2019LanguageMA} pre-trained GPT-2 xl model, and report its perplexity (PPL), as an automatic measure of fluency. Although this measure is not a perfect indicator of fluency, we find it to be a useful metric alongside human judgements.~\footnote{Due to the high variance in the PPL scores generated across sentences by GPT-2, we report the median score for each system under comparison.} 

\noindent\textbf{BLEU.} For sentiment (Yelp) and formality (GYAFC)  transfer where we have reference text, we report the BLEU score. For controlled debiasing, we report BLEU  between generated text and source and show it as BLEU (src).

\noindent\textbf{BertScore.}
As a measure of meaning preservation, we use the F1 BertScore metric~\cite{bertscore} to compare the semantic similarity of the provided reference sentence with the generated output.

\noindent\textbf{Hamming Distance.}
We also report the hamming distance between the source text and generated text, to measure the extent of the change.

\subsubsection{Attribute Quality}
\noindent\textbf{Internal Classifier Accuracy.}
We report the accuracy of the internal classifier (the discriminator used for generation) on the generated text, assuming the target attribute is the correct label. The higher this accuracy is, the better.

\noindent\textbf{External Classifier Accuracy.}
It is natural to get high accuracy on the internal classifier, since we are sampling from it. To have a fair comparison, we  report  accuracy using external classifiers  from Huggingface  (\texttt{textattack/bert-base-uncased-yelp-polarity}~\cite{morris2020textattack} for sentiment and \texttt{cointegrated/roberta-base-formality} for formality).

\noindent\textbf{Agency Lexicon Accuracy.}
 For controlled debiasing, we measure the accuracy of the change in agency by comparing the target agency level with that of the generated text, extracted using the connotation frames lexicon, and following the setup from~\citet{ma-etal-2020-powertransformer}.

\section{Results}

\begin{table*}[]
    \centering
    \caption{Original and style transferred sample sentences, using Mix \& Match LM. Sentiment shows the task of sentiment transfer, from negative to positive and positive to negative, on Yelp. Agency shows the controllable agency de-biaisng task~\cite{ma-etal-2020-powertransformer}. In the examples, we are transferring negative agency to positive.}
    \vspace{-2ex}
    \label{tab:sent_transfer}
    \begin{adjustbox}{width=1.01\linewidth, center}
     \newcolumntype{L}{>{\RaggedLeft\arraybackslash}p{0.06\linewidth}} 
  \newcolumntype{O}{>{\RaggedLeft\arraybackslash}m{0.07\linewidth}} 
  \newcolumntype{D}{>{\arraybackslash}m{0.15\linewidth}} 
  \newcolumntype{R}{>{\arraybackslash}m{0.47\linewidth}} 
\begin{tabular}{@{}l@{\hskip 2mm}p{9.5cm}p{11cm}@{}}
	\toprule
	& {Original} & { Transferred} \\

    \midrule
  
   \multirow{4}{*}{\STAB{Sentiment}}
   & the food 's ok , the service is among the worst i have encountered .
  & the food 's wonderful , the service is among the finest i have encountered .
 \\
   &we will not be using this location again .
 & we will definitely be seeking this location again .
\\

   &good selection of parts and accessories and reasonable prices .
 &poor selection of parts and accessories and high prices .
\\
&it is a cool place , with lots to see and try .
&it is a stupid place , with nothing to see and try .
\\
  
    \midrule
   \multirow{4}{*}{\STAB{Agency}} 
   & mary needed new shoes . 
&	mary got new shoes .
\\
   & 	she followed the instructions as best as she could . 
 & she executed the instructions as best as she could .
\\
& pam wanted to have a special cake for her son 's birthday . 
&pam decides to have a special cake for her son  's birthday .
\\
 &	whitney is going to fail her test . & whitney is set to get her test .
\\

	\bottomrule
\end{tabular}

    \end{adjustbox}
     \vspace{-1ex}
\end{table*}

\subsection{Controllable Debiasing}
\begin{table*}[]
    \centering
    \caption{Controllable debiasing/ sentence agency revision on ROC-story corpus. The \textit{(src)} next to the metrics denotes measurement with respect to the source text. \textit{Int. Clsf.} is the accuracy of the discriminator used in the energy. \textit{Hamm.} shows the Hamming distance. \textit{Agency Acc.} is the accuracy of agency revision based on the agency lexicon (Sec~\ref{sec:metric:lang}).}
    \vspace{-2ex}
    \label{tab:bias}
    \begin{adjustbox}{width=\textwidth, center}
     \newcolumntype{L}{>{\RaggedLeft\arraybackslash}p{0.06\linewidth}} 
  \newcolumntype{O}{>{\RaggedLeft\arraybackslash}m{0.07\linewidth}} 
  \newcolumntype{D}{>{\arraybackslash}m{0.15\linewidth}} 
  \newcolumntype{R}{>{\arraybackslash}m{0.24\linewidth}} 
\begin{tabular}{@{}clcccccc@{}}
	\toprule
	& {Method} & {BLEU(src)} & { GPT-2} &{BertScore(src)}&{Hamm.(src)}	& {Int. Clsf.} & {Agency Acc.} \\

    \midrule
 
   & Source Text   & 100.00& 153.9	&	1.00&	0.00& 7.47&9.81	\\
       \midrule[0.1pt]
       \multirow{2}{*}{\STAB{Basel.}} 
    &\textbf{PowerTransformer (No Boost)}	    & 60.30& 	210.8 &\textbf{0.94} & 1.11 &64.84	& \textbf{69.17}	\\
    &\textbf{PowerTransformer (+Boost) }   &	57.46  & 247.2&	\textbf{0.95}&  1.28&77.23	& \textbf{85.03}	\\
    \midrule[0.1pt]
    \multirow{6}{*}{\STAB{Ours}} 
    &M\&M LM Verb Replace (Disc)	&60.53 & 238.7&0.95 &1.04 &81.05&70.80	\\
    &M\&M LM Verb Replace (Agency Score )	&63.34& 193.3 & 0.96& 0.89& 32.42&64.75	\\
    &M\&M LM Verb Replace (Disc+Agency Score)	&  54.52& 248.8&0.95 & 1.05&77.23&77.27	\\
 
    &\textbf{M\&M LM (Hamming +Disc)}	&56.26 &211.2 & \textbf{0.95}& 1.37&96.52&\textbf{69.00}	\\
    &M\&M LM (Hamming+Agency Score )	&35.26 & 231.6&0.95 &1.56 & 23.13&86.01	\\
    &\textbf{M\&M LM ( Hamming+Disc+Agency score)}	&  39.82&261.6& \textbf{0.93}&2.45 & 90.16& \textbf{89.42}	\\

	\bottomrule
\end{tabular}

    \end{adjustbox}
     \vspace{-1ex}
\end{table*}

Tables~\ref{tab:sent_transfer} and~\ref{tab:bias} show our results for the task of text revision for controlling agency bias which  is introduced by PowerTransformer~\citealt{ma-etal-2020-powertransformer}, our Baseline for this task. 
PowerTransformer has a vanilla (no boost) variant and a variant with vocab boosting, which up-weights the logits of verbs that belong to the target agency lexicon so as to increase their probability and incentivize generation in that direction.  We also measure our metrics on the original test-set, without revision, to provide a better sense of the changes made. 

We offer different variants of our framework, to provide a fair comparison and to better ablate our proposed method. ``Disc'' denotes our framework where we add the discriminator expert ($E_{\texttt{disc}}$) which is trained to predict the agency level of a sentence, to the energy along with $E_{\texttt{mlm}}$, and $E_{\texttt{hamm}}$ (Eq.~\ref{eq:exp}). Hamming distance is computed between the generated proposals and the source sentence. 
The ``Agency Score'' variant adds an alternative term to $E_{\texttt{M\&M}}$ instead of $E_{\texttt{disc}}$, which is the number of target agency verbs according to the connotation frames lexicon~\cite{sap-lexicon} in the sentence. The ``Disc+Agency'' variant has both energy components. We also apply our method in two ways: ``Verb Replace'' which allows the sampler to propose revisions for only one pre-determined verb (provided in the dataset). In this setup, all tokens remain frozen, except for the given verb. The conventional mode (M\&M LM), however, proposes revisions for all tokens in the sentence and is not constrained.

Table~\ref{tab:bias} shows that in the conventional setup, Mix and Match LM (Disc only) has performance similar to that of PowerTransformer, without boosting. With the Agency Score component, our method outperforms PowerTransformer in terms of accuracy of revision as per the agency lexicon accuracy metric, with negligible loss in meaning (BertScore). 
The reason behind this better performance in terms of applying target agency accuracy is that our method's sampling is guided by the energy that is directly built on the metrics we care about, as opposed to trying to apply them through paraphrasing and proxies such as vocab boosting, which are employed in the PowerTransformer method.

Another important observation here is the difference between ``Verb Replace'' and conventional modes. This ablation shows that although our method makes few changes (the average Hamming distance between source and output sentences are between $1.37$ and $2.45$), it still outperforms a ``static'' method that has extra knowledge of the offending verb and focuses on changing only that verb, by a significant margin. 
\subsection{Style Transfer}
In this section we experiment with sentiment and formality transfer, where Sentiment transfer needs fewer changes and formality transfer needs more structural change to the original sentence.  We show sample sentences and transfers in Table~\ref{tab:sent_transfer} (we cannot show samples for formality as the dataset is not public).
\subsubsection{Sentiment Transfer}
\begin{table*}[]
    \centering
    \caption{Sentiment transfer on Yelp. \textit{(ref)/(src)} means the metric measured is measured with respect to reference/source text. \textit{Int./Ext. Clsf.} show  internal/external attribute classifier accuracy. \textit{Hamm.} shows  Hamming distance.}
    \vspace{-2ex}
    \label{tab:sentiment}
    \begin{adjustbox}{width=\textwidth, center}
     \newcolumntype{L}{>{\RaggedLeft\arraybackslash}p{0.06\linewidth}} 
  \newcolumntype{O}{>{\RaggedLeft\arraybackslash}m{0.07\linewidth}} 
  \newcolumntype{D}{>{\arraybackslash}m{0.15\linewidth}} 
  \newcolumntype{R}{>{\arraybackslash}m{0.34\linewidth}} 
\begin{tabular}{@{}cRcccccc@{}}
	\toprule
	& {Method} & {BLEU(ref)} & {GPT-2} &{BertScore(src)}&{Hamm.(src)}	& {Int. Clsf.} & {Ext. Clsf.} \\

    \midrule
	  &Reference Text   & 100.00& 169.5	&1.00&	5.80&	83.70&85.60	\\
    \midrule[0.1pt]
    \multirow{2}{*}{\STAB{Basel.}} 
    &{~\citeauthor{he2020}}	    &	18.67  & 200.6&	{0.93}& 4.23 &84.87	& {79.82}	\\
    &\textbf{UNMT}     &	17.00  & 171.8 &	\textbf{0.94}&3.67 &84.87	& \textbf{80.22}	\\
    \midrule[0.1pt]
    \multirow{2}{*}{\STAB{Ours}} &
    M\&M LM  (Discriminator $\uparrow$)	&15.75 & 163.5 & 0.93& 2.84&97.53&90.00	\\
    &\textbf{M\&M LM (Hamming$\uparrow$)}	&19.71 & 191.5 & \textbf{0.95}& 1.83&94.72&\textbf{82.85}	\\
	\bottomrule
\end{tabular}

    \end{adjustbox}
\end{table*}

\begin{table*}[]
    \centering
    \caption{Formality transfer on GYAFC dataset. The \textit{(ref)/(src)} next to the metrics denotes that they are measured with respect to the reference/source text. \textit{Int. Clsf.} shows the accuracy of the discriminator used in the energy, and \textit{$\rightarrow$Informal/Form.} shows the breakdown of the external classifier accuracy. \textit{Hamm.} shows the Hamming distance.}
    \vspace{-2ex}
    \label{tab:formality}
    \begin{adjustbox}{width=\textwidth, center}
  \newcolumntype{O}{>{\RaggedLeft\arraybackslash}m{0.07\linewidth}} 
  \newcolumntype{D}{>{\arraybackslash}m{0.15\linewidth}} 
  \newcolumntype{R}{>{\arraybackslash}m{0.29\linewidth}} 
\begin{tabular}{@{}cRccccccc@{}}
	\toprule
	& {Method} & {BLEU(ref)} & {GPT-2} &{BertScore(src)}&{Hamm.(src)}	& {Int. Clsf.} & {$\rightarrow$Informal} &{$\rightarrow$Form.} \\

    \midrule
 
    &Reference Text   & 100.00	&  118.1 &0.92&7.72&82.97&	100.00& 9.41	\\
     \midrule[0.1pt]
    \multirow{2}{*}{\STAB{Basel.}} & 
    \textbf{~\citeauthor{he2020}}	    &	15.83  & 122.8 &	\textbf{0.90}&  10.03&64.79	& \textbf{100.00}&\textbf{3.33}	\\
    &UNMT 	    & 14.17	&143.8	&0.90&11.92&56.04&99.81&7.64	\\
    \midrule[0.1pt]
    \multirow{1}{*}{\STAB{Ours}} &
    M\&M LM  (Discriminator $\uparrow$)	&17.78 & 206.3& 0.89& 5.22&91.15&96.67&23.13	\\
    &\textbf{M\&M LM  (BertScore$\uparrow$)}	&27.71 &194.4 &\textbf{0.93}& 2.50& 72.12&\textbf{94.26}&\textbf{19.01}	\\
	\bottomrule
\end{tabular}

    \end{adjustbox}
     \vspace{-1ex}
\end{table*}

For this task, we include two components in our energy model, the attribute discriminator ($E_{\texttt{disc}}$), to induce the target style, and the hamming distance ($E_{\texttt{hamm}}$), to maintain the meaning of the sentence. We don't include the more complex semantic similarity-related component like $E_{\texttt{fuzzy}}$, since sentiment transfer can normally be done by making only a few changes to the sentence.  We report results with two different variants, one where the discriminator component has a higher coefficient in the energy (Discriminator$\uparrow$) and one where the hamming distance has a higher coefficient (Hamming$\uparrow$). In effect, these two show the trade-off between transfer quality and faithfulness to the source sentence.

We see in Table~\ref{tab:sentiment} that our method, with the hamming component up-weighted, outperforms both the generative baselines in terms of transfer accuracy (Ext. Clsf.) and semantic similarity (BertScore). We can also see Mix and Match LM has higher BLEU score, with respect to the provided hand-written reference sentences.  
We hypothesize that this superiority is due to the tendency of our model to make minimal revisions that satisfy the product of experts energy model. Therefore, our model can successfully change the style without changing the meaning of the sentence. The generative baselines, however, regenerate the sentence which imposes more change, as can be observed from the hamming distance column (Hamm.(src)) in Table~\ref{tab:sentiment}.

\subsubsection{Formality Transfer}
For this task, we include the formality classifier ($E_{\texttt{disc}}$), Hamming distance ($E_{\texttt{hamm}}$), and BertScore ($E_{\texttt{fuzzy}}$) components in the energy formulation, to permit the transfer of style and also maintain the meaning of the sentence. $E_{\texttt{fuzzy}}$ helps with imposing semantic similarity between the source and generated sentences, since Hamming alone isn't sufficient for judging comparable formal and informal sentences. We show results for two setups of our framework, one where the discriminator coefficient is higher (Discriminator$\uparrow$) and another where the BertScore coefficient is higher (BertScore$\uparrow$).

In Table~\ref{tab:formality}  we have broken down the external classifier accuracy for the different transfer directions of formal to informal ($\rightarrow$ Inf.) and vice versa. We do this because the $\rightarrow$ Form. task is generally harder and therefore has lower accuracy.
We observe that our method outperforms the baselines in terms of BertScore and BLEU, for similar levels of external classifier accuracy. However,  we can see that the GPT-2 PPL of our method is higher than the baselines.
The reason behind this is the format and noise in the data. The samples for this dataset are taken from the music and entertainment industry domain and contain some symbols and characters similar to emojis (e.g. ``:)'' and ``***''). This is where the tendency of our approach toward minimal revisions is hurtful--our revisions of text, often do not get rid of all of these symbols, while the baselines' generative methods successfully remove all the superfluous characters because they rewrite sentences from scratch.

\begin{table*}[]
    \centering
    \caption{Samples of  prompted sentiment controlled  generations, using our Mix and Match LM and PPLM.}
    \vspace{-2ex}
    \label{tab:gen_pplm}
    \begin{adjustbox}{width=1.01\linewidth, center}
     \newcolumntype{L}{>{\RaggedLeft\arraybackslash}p{0.06\linewidth}} 
  \newcolumntype{O}{>{\RaggedLeft\arraybackslash}m{0.07\linewidth}} 
  \newcolumntype{D}{>{\arraybackslash}m{0.15\linewidth}} 
  \newcolumntype{R}{>{\arraybackslash}m{0.47\linewidth}} 
\begin{tabular}{@{}l@{\hskip 2mm}ll@{}}
	\toprule
	& {Ours (Mix and Match LM)} & { PPLM} \\

    \midrule

   \multirow{4}{*}{\STAB{Pos Sent.}} & the country is noted for attracting a quarter-million tourists. & the country's top cycling event is right behind the olympics, and the
 \\
   & the lake we come across can be said to be beautiful. & the lake is a great spot for swimming, diving and snorke
\\
& the chicken and all the other ingredients produced a delicious meal. & the chicken wing is one of the best foods you can eat and it
\\
& the movie was family-friendly and a success in japan. & the movie, which is currently only the third the the the the the
\\
    \midrule

   \multirow{4}{*}{\STAB{Neg Sent.}} & the country was unstable and was not ready to modernize. & the country's top animal welfare agency, the ministry of agriculture and food
\\
& the lake was not supposed to be navigable under any circumstances. & the lake, a large, and the most massive and most terrible of
 \\
& the chicken was growling and beginning to feel a little sick.
& the chicken noodles are the most horrible food i have ever had.
\\
& the movie received only two nominations and earned no grand prix.   & the movie is not in the  , a, a, a\\

	\bottomrule
\end{tabular}

    \end{adjustbox}
     \vspace{-1ex}
\end{table*}

\begin{table*}[]
    \centering
    \caption{Prompted sentiment controlled generation results and human evaluations.\textit{BERT} denotes the BERT MLM energy score (equivalent of GPT-2 perplexity), and lower score is better.  \textit{Int./Ext. Clsf.} show the accuracy of the discriminator used in the energy/external discriminator from Huggingface.}
    \vspace{-1ex}
    \label{tab:pplm}
    \begin{adjustbox}{width=\linewidth, center}
     \newcolumntype{L}{>{\RaggedLeft\arraybackslash}p{0.06\linewidth}} 
  \newcolumntype{O}{>{\RaggedLeft\arraybackslash}m{0.07\linewidth}} 
  \newcolumntype{D}{>{\arraybackslash}m{0.15\linewidth}} 
  \newcolumntype{R}{>{\arraybackslash}m{0.29\linewidth}} 
\begin{tabular}{@{}lSSSSSSSSSSSSSS@{}}
	\toprule
	 {\multirow{2}{*}{Length}}  &\multicolumn{2}{c}{GPT-2~($\downarrow$)}	&   {}& \multicolumn{2}{c}{BERT~($\downarrow$)}	& {}& \multicolumn{2}{c}{Int. Clsf.~($\uparrow$)} &{}& \multicolumn{2}{c}{Ext. Clsf.~($\uparrow$)} &{}&\multicolumn{2}{c}{Human Preference (\%)} \\
	\cmidrule{2-3} \cmidrule{5-6} \cmidrule{8-9} \cmidrule{11-12} \cmidrule{14-15}  
	&                                  {Ours} &  {PPLM}          &                   &     {Ours} &  {PPLM}    & &     {Ours} &  {PPLM}    & &    {Ours} &  {PPLM}    &&    {Ours} &  {PPLM}     \\
    \midrule
    12    &  264.1&113.1& &    -160.4&-137.1&&	 94.3&71.7 && 65.1&58.0 &&71.1 & 29.9\\
    20  &	167.2&61.1& & -271.0&-237.1 &&  96.3&74.5 && 65.9&57.6 && 62.9 & 37.1\\
	50 &   122.3&29.0& &  -692.3&-606.1&& 93.8 & 73.6 && 68.6 &60.7 && 46.7 & 53.3\\
	\bottomrule
\end{tabular}


    \end{adjustbox}
\end{table*}

\subsection{Prompted Controlled Generation}

\subsubsection{Sentiment Controlled Generation}
We generate 560 sequences of different lengths ($12$, $20$ and $50$ tokens),  given 14 prompts, 2 sentiments, and 20 sequences per sentiment, taken from~\citet{pplm}'s experimental setup. The prompts and sample generations are in the appendix~\ref{app:prompts} and~\ref{app:generation}, and a full list of generations is in the supplementary material. 

Table~\ref{tab:pplm} shows our results for this experiment. Here, we have an additional metric, the MLM energy (lower is better), which, like GPT-2, indicates the quality of generated sentences~\cite{salazar-etal-2020-masked} according to BERT. We report this extra metric here since PPLM uses a GPT model for generation, and it is natural that it would measure better on this metric. 
The table shows that for all lengths of generated sentences, our method is much better at inducing the target sentiment. 
However, we observe that PPLM performs better in terms of GPT-2 while our method performs better on the MLM energy metric. This suggests the tendency of model-based fluency metrics to be biased toward the corresponding models as the PPLM uses GPT-2 for generation and M$\&$M LM uses BERT.
%
To enable a more conclusive comparison of the text quality, we report results with human evaluations. For these evaluations, we randomly select 10 generated outputs for each prompt, per sentiment (240 overall), and asked three Amazon Turkers per sample pair,  which sample they find more fluent. We report the majority vote of the Turkers in the table. The results show that for sequences with lengths 12 and 20, they found our generations more fluent. However, for length 50, the preference rate for M\&M drops to $46.7\%$, which shows that our method is superior to PPLM for short/medium length generation, however, PPLM does better at generating longer sequences.

\subsubsection{Topic Controlled Generation}
We follow FUDGE's~\cite{fudge} experimental setup which covers $7$ topics, given $20$ prompts and generate $7 \times 20$ sequences of length $20$.
To enforce topicality on our generations, we add a topic-based energy, $E_{\texttt{topic}}$. This energy is essentially the negative count of the number of topic-related words (using the list provided by FUDGE). 

Table~\ref{tab:fudge} shows the results of this experiment, generations are also provided in~\ref{app:generation}. Topic-score~($\uparrow$) is the usage rate of topic-related words that were used for training and evaluation of topic controlled generation by~\citeauthor{fudge} in their paper. 
Grammaticality~($\uparrow$) is the score of grammaticality given by a Roberta-based CoLA grammaticality model averaged over all outputs~\cite{warstadt2019neural}. 
The ``Div''~($\uparrow$) metrics show the diversity of generated text, over unigrams, bigrams and trigrams.
Finally, the human evaluations show human preference, in terms of fluency of the sentences (~\ref{app:human}).  
As shown by the table, the fluency of our method is comparable to that of FUDGE, even better in terms of human preference and grammaticality judgment. 
FUDGE has a slightly higher topic score, which is expected since it trains a custom step-wise discriminator for each topic that is optimized for the task. But our approach shows competitive faithfulness to the topics especially considering the fact that prompted GPT-2 generations without the FUDGE discriminators only achieve a topic-score of $0.23$.

\subsection{Inference Speed}

Given that our model's inference procedure involves MCMC sampling, it's reasonable to expect its run-time to be slower than more traditional baselines. For sequences of length 20, we find that our un-optimized implementation requires 8 seconds per generation and 3 seconds per revision -- while, in contrast, baseline system PPLM requires 16 seconds and FUDGE requires 0.4 seconds per generation. This is a substantial slowdown compared to FUDGE, but not one that renders the proposed approach impractical in offline settings. Further, faster sampling schemes are beyond the scope of this paper but might be explored in future work to speed up models like M\&M LM. 
\begin{table}[]
    \centering
    \footnotesize
    \caption{Prompted topic controlled generation results and human evaluations. 
    }
    \vspace{-2ex}
    \label{tab:fudge}
  \newcolumntype{O}{>{\RaggedLeft\arraybackslash}m{0.07\linewidth}} 
  \newcolumntype{D}{>{\arraybackslash}m{0.15\linewidth}} 
  \newcolumntype{R}{>{\arraybackslash}m{0.29\linewidth}} 
\begin{tabular}{lrr}
	\toprule
	Metrics & FUDGE & \textbf{M\&M LM} \\
	\midrule
	Topic-score~($\uparrow$) & 1.45 & 1.21 \\
	Grammaticality~($\uparrow$) & 0.61 & 0.74\\
	GPT-2 PPL~($\downarrow$) & 104.8 & 110.2\\
	Diversity over Unigrams~($\uparrow$) &  0.54& 0.57\\
	Diversity over Bigrams~($\uparrow$) & 0.86 & 0.89\\
	Diversity over Trigrams~($\uparrow$) & 0.87 & 0.88\\
	Human Preference(\%)~($\uparrow$) & 36.5 & 63.5\\



	\bottomrule
\end{tabular}

    \vspace{-1ex}
\end{table}

\section{Conclusion}
We present Mix and Match Language Models (M\&M LM), a training-free framework for controlled text generation that can easily mix heterogeneous expert modules. We show that our framework outperforms prior methods on a suite of text revision and attribute-controlled generation tasks. Further, our results indicate that probabilistic energy language models, typically considered intractable, can be used for practical text generation tasks when combined with an appropriate sampling scheme.

\section*{Acknowledgments}
The authors would like to thank the anonymous reviewers and meta-reviewers for their helpful feedback. We also thank our colleagues at the UCSD/CMU Berg Lab for their helpful comments and feedback. 

\section*{Ethical Considerations} 

The proposed approach takes steps towards a novel paradigm that might partially mitigate the need for energy-intensive GPU training -- potentially leading to positive environmental impact down the line. The approach may also have positive impacts on accessibility as strong computational resources are not required when setting up a new controlled text generation system. 
%
%
We do however acknowledge that strong controlled generation methods that rely on discriminators have the potential to regurgitate sensitive training data and produce harmful outputs and toxic language~\cite{xudetoxifying,gehman2020realtoxicityprompts,Wallace2020ImitationAA}. However, if used properly and for good, we anticipate a positive impact on debiasing and safe generation.

\bibliography{acl_latex.bib}
\bibliographystyle{acl_natbib}

\appendix
\clearpage

\begin{table*}[]
    \centering
    \caption{Samples of  prompted topic controlled  generations, using our Mix and Match LM and FUDGE.}
    \vspace{-2ex}
    \label{tab:gen_fudge}
    \begin{adjustbox}{width=1.01\linewidth, center}
     \newcolumntype{L}{>{\RaggedLeft\arraybackslash}p{0.06\linewidth}} 
  \newcolumntype{O}{>{\RaggedLeft\arraybackslash}m{0.07\linewidth}} 
  \newcolumntype{D}{>{\arraybackslash}m{0.15\linewidth}} 
  \newcolumntype{R}{>{\arraybackslash}m{0.47\linewidth}} 
\begin{tabular}{@{}l@{\hskip 2mm}p{11cm}p{11cm}@{}}
	\toprule
	& {Ours (Mix and Match LM)} & { FUDGE} \\

    \midrule

   \multirow{3}{*}{\STAB{Computer}} 
 &	to review, please link to (chessworld.net/chessworld/download.html). &  to review, instead of using the "n/a" flag (like on our previous posts) 
\\
   & in summary, key program clients are homeforge, blogdev and skeptic.net. &	in summary:- install and run a local mysql server on the host computer- add a mysql table 
\\
   & 	it has been shown using several techniques, including microscopy, electron microscopy, and digital loansharking. &it has been shown using ebpf/ebpis (extraction of a new ebp
\\

    \midrule

   \multirow{3}{*}{\STAB{Legal}}
   
   &the connection to the assault was not without controversy, especially given the expert testimony the prosecutor had provided. &  the connection failed, however, under an audit of one of the two, the judge said. the	
\\
& to review, or submit information to the cdu regarding the current (constitutionally) electoral law.  & to review, the court's decision not to review the case raises an important question. the court's	
 \\
   & to conclude, when a claim is not true, the defendant's claims are often not true. &	to conclude, the court held a motion is properly made to dismiss a claim for an award of attorney
\\

\midrule
   \multirow{3}{*}{\STAB{Military}}
   &foundational to this is the cold war, which eliminates all military defense available to the enemy. &	foundational to this is an attack on the conventional wisdom on the left that the left is the party
\\
& views on the civil war fleet, the national maritime museum. views on the royal navy, admiralty.& 	views on russia's military buildup on the strength of his repeated insistence, a number of
 \\
&to conclude, we all agree that constructive defense methods are not yet available.& constructive defense? 	to conclude, the russian navy's attack on the malaysian ship, a taskforce carrying out exercises,
\\

\midrule
   \multirow{3}{*}{\STAB{Politics}}
   &an illustration of: the historical background, culture, and general political significance of the books' contents. &	an illustration of an anti-democratic regime under a fascist dictatorship and its suppression of the popular opposition and
\\
& the issue focused on socialism, democracy, social justice and self-government in countries across the globe. & 	the issue focused on religious freedom in the country's constitution, a fundamental pillar of u.s.
 \\
&	in this essay, king frederick iii of prussia was prominently featured in american post-civil war culture.  & in this essay, the term "political correctness" is used to refer to political demands imposed on the
\\

\midrule
   \multirow{3}{*}{\STAB{Religion}}
  &the issue focused on the inferiority of conservatives ( "" religious conservatives "" ) vs . atheists . &	the issue focused on religious freedom, particularly when the bible teaches that god is "the creator." 
\\
&to summarise accurately the wind direction, additional characters may be added to the classification table below. &	to summarise, if the present-day christian churches were a monastic order of the monks rather
 \\
& 	an illustration of the natural history of wales by francis bacon. bateson, charles (1839).  &an illustration of an ancient bronze age village in the northern greek region of crete, which shows a
\\

\midrule
   \multirow{3}{*}{\STAB{Science}}
  &prior to this date, the manuscript was not currently available on the internet, and is cited rarely. &	prior to this experiment, the scientists had not seen a new species in the area since the late 1800

\\
&the relationship has inspired research into the role of women in economics, and contributions to feminist economic theory. 
 &the relationship between energy use and energy use as a function of time was also investigated using a linear mixed	
 \\
& 	the issue focused on developments in the field of "darwinism, biology and human evolution" research. &	the issue focused on data retention, and the key elements of the retention matrix, including retention of identifiers
\\
\midrule
   \multirow{3}{*}{\STAB{Space}}
  &furthermore, the performance space is "packed with classical music" and is "lavishly decorated". &	furthermore, the eighty-first star is the planet's largest moon and it sits directly in between

\\
&to conclude, an asteroid becomes, mathematically, the largest asteroid to ever be "discovered". 	&to conclude, scientists behind spacemonkey, and a number of the other projects that nasa is supporting
	
 \\
& to summarise other countries'respective territorial claims, including territorial waters, islands, etc. . &	to summarise: x (1x a2 a19 a1 a2 b2

\\
	\bottomrule
\end{tabular}

    \end{adjustbox}
     \vspace{-2ex}
\end{table*}

\begin{table*}[ht!]
    \centering
    \caption{Sentiment transfer on Yelp dataset ablation study. The tuples in the first column show the $(\alpha,\delta,\beta)$ set of parameters. We ablate the effect that different components have on the transfer.The \textit{(ref)/(src)} next to the metrics denotes that they are measured with respect to the reference/source text. \textit{Int./Ext. Clsf.} show the accuracy of the discriminator used in the energy/external discriminator from Huggingface. \textit{Hamm.} shows the Hamming distance.}
    \vspace{-2ex}
    \label{tab:abl_sent}
    \begin{adjustbox}{width=\textwidth, center}
     \newcolumntype{L}{>{\RaggedLeft\arraybackslash}p{0.06\linewidth}} 
  \newcolumntype{O}{>{\RaggedLeft\arraybackslash}m{0.07\linewidth}} 
  \newcolumntype{D}{>{\arraybackslash}m{0.15\linewidth}} 
  \newcolumntype{R}{>{\arraybackslash}m{0.29\linewidth}} 
\begin{tabular}{@{}RSSSSScSSSS@{}}
	\toprule
	 {(Disc, MLM, Hamm.)} & {BLEU} & {GPT-2} &{BertScore}&{Hamm.}	& {Int. Clsf.} & {Ext. Clsf.} \\

    \midrule
  
    $(1,0,1)$     &   4.77   &   	1611.8   &	    0.88	&   5.308   &	81.7	&   67.4	\\
    $(1,0,0)$	    &   1.12    &   	3825.3   &	0.85	&   8.378   &99.0	&   84.5	\\
   $(0,1,0)$	    &   3.77    &   	101.3   &	0.90	&   5.92	&   24.7    &	29.3	\\
     $(100,1,0)$	&   2.89    &   	143.0   &	0.88	&   7.067   &99.2	&   96.5	\\
     $(0,1,50)$	&   23.60	    &     110.0   &	0.99	&   0.002   &	4.3 &	5.0	\\
    $(100,1,50)$	    &   19.71    &   	191.5   &	0.95	&   1.838   &	94.7	&82.8	\\
	\bottomrule
\end{tabular}

    \end{adjustbox}
\end{table*}

\begin{table*}[h]
    \centering
    \caption{Formality transfer on GYAFC dataset ablation study. The tuples in the first column show the $(\gamma,\eta)$ set of parameters. We ablate the effect the BLEURT and BertScore experts have on the transfer. The \textit{(ref)/(src)} next to the metrics denotes that they are measured with respect to the reference/source text. \textit{Int. Clsf.} shows the accuracy of the discriminator used in the energy, and \textit{$\rightarrow$Informal/Form.} shows the breakdown of the external classifier accuracy. \textit{Hamm.} shows the Hamming distance.}
    \vspace{-2ex}
    \label{tab:abl_form}
    \begin{adjustbox}{width=\textwidth, center}
     \newcolumntype{L}{>{\RaggedLeft\arraybackslash}p{0.06\linewidth}} 
  \newcolumntype{O}{>{\RaggedLeft\arraybackslash}m{0.07\linewidth}} 
  \newcolumntype{D}{>{\arraybackslash}m{0.15\linewidth}} 
  \newcolumntype{R}{>{\arraybackslash}m{0.29\linewidth}} 
\begin{tabular}{@{}RSSSSScSSSS@{}}
	\toprule
	 {(BLEURT,BertScore)} & {BLEU} & {GPT-2} &{BertScore}&{Hamm.}	& {Int. Clsf.} & {$\rightarrow$Inf.} &{$\rightarrow$Form.} \\

    \midrule
  
    $(100,0)$    & 14.07&	243.9&0.87&5.93	&89.34	&97.41&19.80	\\
    $(300,0)$	       & 13.75&	233.9&0.88&5.88	&89.34	&97.01&22.94	\\
   $(0,100)$	    & 17.78&	206.3&0.89&5.22	&91.15	&96.67&23.13	\\
     $(0,300)$	   & 18.85&	210.9&0.90&4.91	&88.23	&97.04&23.13	\\
 
	\bottomrule
\end{tabular}

    \end{adjustbox}
\end{table*}

\section{Appendix}
\label{sec:appendix}

\subsection{Code and Data Directory Structure}\label{app:code}
We have provided all our code, data and our generations in~\url{https://github.com/mireshghallah/mixmatch}, and our checkpoints are uploaded anonymously here \url{https://zenodo.org/record/5855005}.  There is a \texttt{readme} file  in the repo, which has instructions on how to run generation and get evaluation metrics. 
%
We have not included the data files for the formality, since the GYAFC dataset requires permission for access, so we cannot release it. 


\subsection{Sample Generations}\label{app:generation}
Due to page limitations in the body of the paper, we include more sample generations from our method in the form of tables here. We have no samples from the formality transfer task, however, since the data used (GYAFC) is protected and needs permissions for access, so we cannot publish it. However, we have provided code needed to reproduce our results, once access to the original data is gained. 
Table~\ref{tab:gen_fudge} shows  FUDGE generations versus Mix and Match generations.

\section{Experimental Setup Details}\label{app:exps}
\subsection{Tasks and Datasets}

\noindent\textbf{Controllable debiasing (ROC story corpus):} We use the subset of the ROC story corpus~\cite{mostafazadeh-etal-2016-corpus} test-set that is used by PowerTransformer~\cite{ma-etal-2020-powertransformer} for their evaluations. We use this data for  controllable debiasing, a text revision task which aims to correct the implicit and potentially undesirable agency biases in character portrayals. This test-set consists of 549 sentences, where 224 sentences have low agency verbs (such as wish, dream, etc.) and the rest have high agency (like pursue, achieve, etc.). The task is to revise the sentences such that the meaning is preserved, but the agency of the sentence is changed in the target direction.  

\noindent\textbf{Sentiment transfer (Yelp):} We use Yelp~\cite{shen2017style} dataset's test-set for the task of sentiment transfer. The test set comprises of 1000 sentences, half with positive and half with negative sentiment. We also have a reference set of hand written sentiment transferred sentences, provided by~\cite{he2020} that we use for reporting evaluation metrics.

\noindent\textbf{Formality transfer (GYAFC):}  We use 1051 sentences from the test-set of the GYAFC~\cite{Rao2018DearSO} dataset, which contains formal and informal sentences for the task of formality transfer (both directions of formal to informal and informal to formal). Here we use the entertainment and music domain subset of this data, following the evaluation setup of~\cite{he2020}. This dataset also contains parallel data between formal and informal sentences, which we use as reference for reporting evaluation metrics.

\noindent\textbf{Prompted generation:} We evaluate our approach on two forms of prompted generation: 1) sentiment controlled generation, and 2) topic controlled generation. on prompted generation. For sentiment controlled generation, we set Mix and Match LM to generate text with positive or negative sentiment given prompts (listed in Appendix~\ref{app:prompts}) by using a Yelp sentiment classifier as discriminator and compare against PPLM~\cite{pplm} which is a popular sentiment controlled generation method. For topic controlled generation, we compare against FUDGE~\cite{fudge}, and follow their experimental setup consisting of 7 distinct topics and 20 prompts. 

\subsection{Expert Component Configurations}

We use a Huggingface pre-trained \texttt{bert-base-uncased} model as our MLM for yielding $E_{\texttt{mlm}}$ and also providing the proposal distribution in our MH MCMC sampler. For obtaining $E_{\texttt{disc}}$, we train BERT-based classifiers on the training-set of our datasets to use as our attribute discriminators. 
Although we could have used any pre-trained attribute classifier from a model repository like Huggingface for $E_{\texttt{disc}}$, we train our own classifier for controlled empirical comparison. As described later, we do use pretrained Huggingface attribute classifiers as external attribute classifiers for fair evaluation against baselines. 
For experiments in which we add the BertScore~\cite{bertscore} component to the energy, we download the pre-trained \texttt{roberta-large\_L17} models from Huggingface, respectively.
We have provided implementation details and hyperparameter ablations of all the experiments in Appendix~\ref{app:bias},~\ref{app:sentiment},~\ref{app:formality} and~\ref{app:prompts}.

\subsection{Baselines}

\noindent\textbf{PowerTransformer.}
For the task of controllable debiasing (agency revision), we compare our work with PowerTransformer~\cite{ma-etal-2020-powertransformer}, an approach that uses paraphrasing and self-supervision based on a reconstruction loss, building on pre-trained language models, to re-write text and control agency level of sentences.

\noindent\textbf{~\citeauthor{he2020}} For style transfer on sentiment an formality domains, we compare our work with~\citet{he2020}, a generative style transfer framework which uses a variational autoencoder (VAE) built using a sequence-to-sequence LSTM-based model to do unsupervised style transfer. This framework needs to be trained from scratch for each style transfer task.

\noindent\textbf{UNMT.} As a second baseline for style transfer, we compare our work with UNMT~\cite{lample2018phrase}, an unsupervised machine translation framework that demonstrates high performance for sentiment transfer.

\noindent\textbf{PPLM.} For the task of sentiment controlled generation, we compare our work to Plug-and-Play LM (PPLM)~\citet{pplm}, which does attribute controlled generation using the flow of gradients from discriminators trained on the last hidden layer representations of the generator, to guide generation. 

\noindent\textbf{FUDGE.} This approach~\cite{fudge} trains step-wise discriminators on partial generations from GPT-2 to determine whether the constraints related to desired attributes will be satisfied by the future completion of the sequence or not. We compare against this on topic controlled generation as this approach was shown to be superior to PPLM on this task.
\subsection{Evaluation Metrics}
We use a variety of evaluation metrics to compare our approach's performance on two major facets: (1) Quality of generated text, and (2) success on matching the target attribute used for control.

\subsubsection{Text Quality and Semantic Similarity}
\label{sec:metric:lang}
\noindent\textbf{GPT-2 PPL.}  We feed our generated test sentences to a Huggingface~\cite{Radford2019LanguageMA} pre-trained GPT-2 xl model, and report its perplexity (PPL), as an automatic measure of fluency. Although this measure is not a perfect indicator of fluency, we find it to be a useful metric alongside human judgements.~\footnote{Due to the high variance in the PPL scores generated across sentences by GPT-2, we report the median score for each system under comparison.} 

\noindent\textbf{BLEU.} For sentiment (Yelp) and formality (GYAFC)  transfer experiments, since we have reference text, we report the BLEU score. For controlled debiasing, we report BLEU  between  generated text and source, and show it as BLEU (src).

\noindent\textbf{BertScore.}
As a measure of meaning preservation, we use the F1 BertScore metric~\cite{bertscore} to compare the semantic similarity of the provided reference sentence with the generated output.

\noindent\textbf{Hamming Distance.}
We also report the hamming distance between the source text and generated text, to measure the extent of the change induced by our framework.

\subsubsection{Attribute Quality}
\noindent\textbf{Internal Classifier Accuracy.}
To evaluate the quality of applying target attributes, we report accuracy of the internal classifier (the discriminator used for generation) on the generated text, assuming the target attribute is the correct label. The higher this accuracy is, the better.

\noindent\textbf{External Classifier Accuracy.}
Since the internal classifier is the one we are sampling from, it is natural that we would get high accuracy on it, compared to our baselines. To create a more fair comparison, we also report classification accuracy using external classifiers, downloaded from Huggingface. For sentiment classification we use \texttt{textattack/bert-base-uncased-yelp-polarity}~\cite{morris2020textattack}, and for formality we use \texttt{cointegrated/roberta-base-formality}.

\noindent\textbf{Agency Lexicon Accuracy.}
 For the controlled debiasing experiment, we measure the accuracy of the change in agency by comparing the target agency level with that of the generated text, extracted using the connotation frames lexicon, and following the setup from~\citet{ma-etal-2020-powertransformer}.
 
\subsection{Hyper-parameter and Component Selection}
Selection of components is based on the needs of the task and is  straight forward. You add each component you need, to satisfy some condition. If you want to do sentiment controlled generation, you add a sentiment classifier. 
Finding the hyperparameters for each component (the multiplier in energy) is also simple, since the trade-off between the different components is clear. For instance, as shown in Table~\ref{tab:abl_sent}, increasing the discriminator score results in a more successful sentiment transfer, and increasing the Hamming score results in keeping the sentence the same.

\subsection{Controllable Debiasing: \\Hyper parameters}
\label{app:bias}
For the results presented in Table~\ref{tab:bias}, we ran the Gibbs chain for 8 epochs (8 iterations over all the tokens) for the conventional mode of our method, and 30 iterations for verb replacement. We used the parameters $\alpha=100,\beta=50,\theta=100$, where $\theta$ is the coefficient assigned to the agency scorer, and  $\alpha$ and $\beta$ are defined in Equations~\ref{eq:disc} and~\ref{eq:exp}.

\subsection{Sentiment Transfer: Hyperparameters}
\label{app:sentiment}
In this section we discuss the hyperparameters used for sampling and see the effects of each one. For the results presented in Table~\ref{tab:sentiment}, we ran the Gibbs chain for 8 epochs (8 iterations over all the tokens), and used the parameters $\alpha=100,\beta=25$ (for Discriminator~$\uparrow$) and $\alpha=100,\beta=50$, for Hamming~$\uparrow$. $\alpha$ and $\beta$ are defined in Equations~\ref{eq:disc} and~\ref{eq:exp}.

Table~\ref{tab:abl_sent}  shows six different scenarios, with six different coefficeints for the Disciriminator ($\alpha$), BERT MLM ($\delta$) and Hamming distance ($\beta$) components in the energy function, which helps understand the effect each expert has.

\subsection{Formality Transfer:  Hyperparameters}
\label{app:formality}
For the results presented in Table~\ref{tab:formality}, we ran the Gibbs chain for 5 epochs (5 iterations over all the tokens), and used the parameters $\alpha=140,\beta=15,\gamma=100$ (for Discriminator~$\uparrow$) and $\alpha=140,\beta=50,\gamma=300$, for BertScore~$\uparrow$. $\alpha$, $\beta$ and $\gamma$ are defined in Equations~\ref{eq:disc} and~\ref{eq:exp}.

Table~\ref{tab:abl_form}  shows four different scenarios, with four different coefficeints for the BLEURT and BertScore components in the energy function, which helps understand the effect each expert has. For BLEURT, we use  pre-trained \texttt{Elron/bleurt-base-512} from Huggingface.

\subsection{Prompts and Hyperparameters Used for Controlled Generation}
\label{app:prompts}
We have listed the prompts that we used for controlled text generation (these prompts are taken from~\citet{pplm}): the country, the lake, the chicken, the movie, the pizza, the painting, the year, the city, the book, the potato, the horse, the road, the president, once upon a time. We collect these prompts from PPLMs github repo, available at this url: \url{https://github.com/uber-research/PPLM/tree/master/human_annotation/pplm_labeled_csvs}.

PPLM has multiple knobs to tune for sampling, and after running a greed search we found that \texttt{gamma=1, num\_iterations=10 , step\_size=0.1, kl\_scale=0.01} and \texttt{gm\_scale=0.95} yeild the best results (reported in Table~\ref{tab:pplm}).  We generated samples by running the command \texttt{python\quad run\_pplm.py \quad-D \quad sentiment}, with the mentioned hyperparameters.
For FUDGE, we tune the $\lambda$ parameter, and we find that $\lambda=10$ works best.

For our method, we ran the Gibbs chain for 15 epochs, and used hyperparameter $\alpha=40$, from Eq.~\ref{eq:disc}. We don't use any experts other than the yelp sentiment classifier, so we don't have any other hyperparamters. 

\subsection{Human Evaluations}\label{app:human}

We used Amazon Mechanical Turk for our evaluations, where each HIT was a two choice question of ``which sentence is more fluent?'' and the providers were paid $\$0.1$ per HIT. We selected Turkers from English speaking countries. We also had each each question answered 3 times (by 3 Turkers), to create redundancy and robustness. 

\subsection{GPU Hours and Infrastructure}

One of the main purposes of this work is to introduce a paradigm in which we re-use existing models and do not retrain. As such, we did not need GPUs for training (we finetuned two  classifier for demonstration purposes, which took less than two GPU hours).

However, we do use GPUs for inference (less computationally intensive), for generating samples. We used an in-house 4GPU server (NVIDIA RTX2080), and the samplings and hyperparameter tuning took an overall of around 10-14 full days on the 4 GPUs. 


\end{document}